\pdfoutput=1
\documentclass[11pt]{article}

\usepackage{acl}

\usepackage{times}
\usepackage{latexsym}

\usepackage[T1]{fontenc}
\usepackage[utf8]{inputenc}

\usepackage{microtype}

\usepackage{inconsolata}

\usepackage{url}            % simple URL typesetting
\usepackage{hyperref}       % hyperlinks
\usepackage{booktabs}       % professional-quality tables
\usepackage{amsfonts}       % blackboard math symbols
\usepackage{nicefrac}       % compact symbols for 1/2, etc.
\usepackage{microtype}      % microtypography
\usepackage{xcolor}         % colors
\usepackage{enumitem}
\usepackage{graphicx}

\usepackage{titletoc}

\newcommand\benchmarkName{\textbf{iSign}}
\newcommand\iSignCount{$118,228$\ }
\newcommand\iSignCountShort{$118$k\ }

\title{iSign: A Benchmark for Indian Sign Language Processing}

\author{{\bf Abhinav Joshi}$^\dagger$ \qquad
{\bf Romit Mohanty}$^\dagger$ \qquad 
{\bf Mounika Kanakanti}$^\mathparagraph$\footnotemark[2] \\ 
\textbf{Andesha Mangla}$^\star$ \qquad \textbf{Sudeep Choudhary}$^\diamond$ \qquad \textbf{Monali Barbate}$^\diamond$ \\ {\bf Ashutosh Modi}$^\dagger$  
 \\ 
        $^\dagger$IIT Kanpur \\%\qquad
        $^\mathparagraph$Max Planck Institute for Psycholinguistics \qquad 
        $^\star$ISLRTC \qquad 
        $^\diamond$Microsoft IDC India \\
  \texttt{\{ajoshi, ashutoshm\}@cse.iitk.ac.in}  
}

\begin{document}
\maketitle
% \begin{abstract}
% This document is a supplement to the general instructions for *ACL authors. It contains instructions for using the \LaTeX{} style files for ACL conferences. 
% The document itself conforms to its own specifications, and is therefore an example of what your manuscript should look like.
% These instructions should be used both for papers submitted for review and for final versions of accepted papers.
% \end{abstract}

\begin{abstract}
Indian Sign Language has limited resources for developing machine learning and data-driven approaches for automated language processing. Though text/audio-based language processing techniques have shown colossal research interest and tremendous improvements in the last few years, Sign Languages still need to catch up due to the need for more resources. To bridge this gap, in this work, we propose \textbf{iSign}: a benchmark for Indian Sign Language (ISL) Processing. We make three primary contributions to this work. First, we release one of the largest ISL-English datasets with more than \iSignCountShort video-sentence/phrase pairs. To the best of our knowledge, it is the largest sign language dataset available for ISL. Second, we propose multiple NLP-specific tasks (including SignVideo2Text, SignPose2Text, Text2Pose, Word Prediction, and Sign Semantics) and benchmark them with the baseline models for easier access to the research community. Third, we provide detailed insights into the proposed benchmarks with a few linguistic insights into the workings of ISL. We streamline the evaluation of Sign Language processing, addressing the gaps in the NLP research community for Sign Languages. We release the dataset, tasks, and models via the following website: \url{https://exploration-lab.github.io/iSign/}. 
\end{abstract}

% \begin{abstract}
%   The abstract paragraph should be indented \nicefrac{1}{2}~inch (3~picas) on
%   both the left- and right-hand margins. Use 10~point type, with a vertical
%   spacing (leading) of 11~points.  The word \textbf{Abstract} must be centered,
%   bold, and in point size 12. Two line spaces precede the abstract. The abstract
%   must be limited to one paragraph.
% \end{abstract}

% \vspace{-4mm}
\section{Introduction} \label{sec:intro}
% \vspace{-2mm}

As per the WHO estimate, about 63 million people belong to the Deaf and Hard of Hearing (DHH) community in India \cite{WHO, varshney2016deafness}. Consequently, Indian Sign Language (ISL) is widely used in the Indian subcontinent. Moreover, according to Ethnologue (\citeyear{ethnalogue-wiki:1092421532}) (a reference publication documenting information about living languages of the world), ISL is the world's most widely used sign language. However, there is a considerable deficit of sign language interpreters, e.g., according to the Government of India organization Indian Sign Language Research and Training Center (ISLRTC) (\url{https://islrtc.nic.in/}), there are only 300 certified sign language interpreters in India. NLP technologies can help in this case. 

\begin{figure}[t]
\centering
 \includegraphics[width=\linewidth]{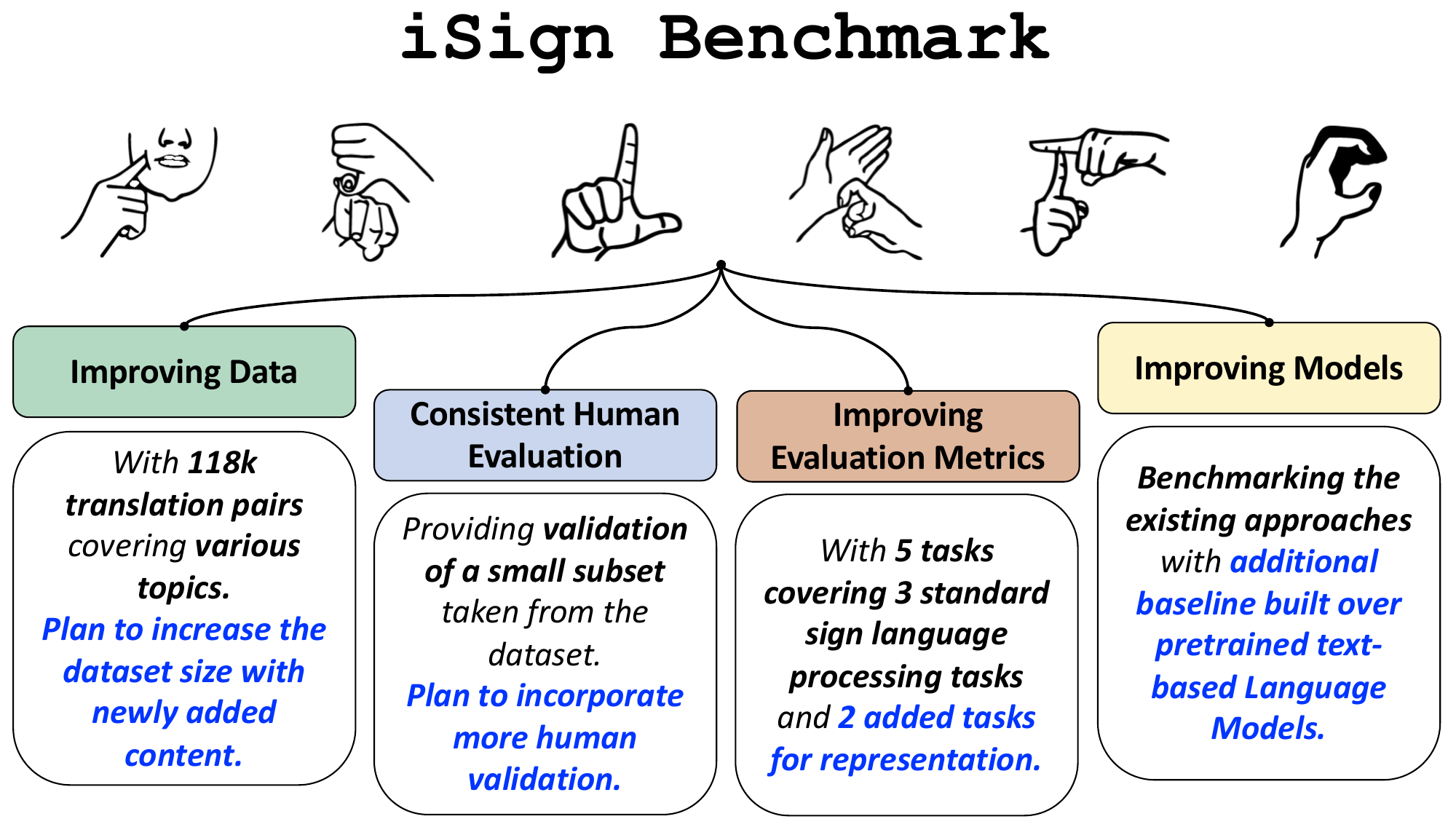}
  \caption{iSign Benchmark: The proposed benchmark for Indian Sign Language Processing. 
  % Text in the image inspired from \cite{gehrmann-etal-2021-gem}
  }
  \label{fig:iSign_thumbnail_intro}
  % \vspace{-6mm}
\end{figure}

\begin{figure*}[t]
\centering
 \includegraphics[scale=0.13]{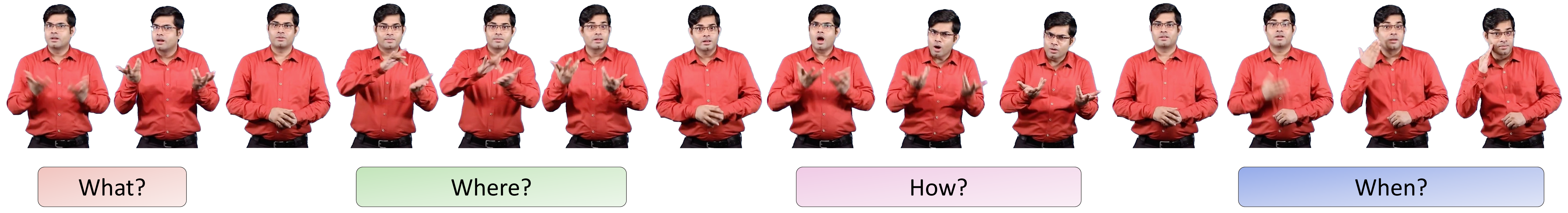}
  \caption{An example showing the translation of the phrase ``What, Where, How, and When" in Indian Sign Language. The text box length overlaps with the signs with a pause position in between.}
  \label{fig:iSign_thumbnail}
  % \vspace{-5mm}
\end{figure*}

\noindent Similar to spoken languages, sign languages are {region-specific}, for example, {people in North America use} American Sign Language (ASL), and {people in Germany use} Deutsche Gebärdensprache (DGS). Considerable efforts have been made to develop technologies for automatically processing sign language in other countries. However, when it comes to ISL, very limited technological advancements have been made; for example, there is a lack of standard benchmarks for ISL, resulting in low development and a lack of comparison of Machine Learning (ML) based solutions for ISL, e.g., word recognition, translation, generation, etc. In contrast, relatively speaking, other sign languages (e.g., American Sign Language (ASL), Deutsche Geb\"ardensprache (DGS)) have a sufficient number of annotated resources for data-driven approaches (Table \ref{tab:translation-comparison}). Natural Language Processing (NLP) has made rapid progress in the last few years \citep{min2021recent}. Most of these approaches have targeted textual datasets. Easy access to textual datasets and leaderboards in languages like English has facilitated the development, reliability, and standardization of experimentation on several tasks \citep{wang2019superglue}. However, there has been limited progress in visual modality-based languages, like sign languages (also referred to as signed languages: \url{https://en.wikipedia.org/wiki/Sign_language}), due to the limited availability of large-scale datasets. Moreover, from a modeling perspective, sign languages are data-hungry due to the complex relationship between different entities in visual modalities like signs, gestures, finger-spelling, and facial expressions. In this paper, to address the lack of a large-scale dataset for ISL processing and to promote the development of sign language processing techniques, we propose \benchmarkName. In a nutshell, we make the following contributions:
\begin{itemize}[nosep,noitemsep]
    \item We introduce \benchmarkName, a new benchmark for Indian Sign Language processing. Figure \ref{fig:iSign_thumbnail_intro} provides an overview and design philosophy of the benchmark (inspired from \cite{gehrmann-etal-2021-gem}).  
    %(A small sample of the dataset is provided in the supplementary material.) 
    \item  We create a dataset with \iSignCount ISL-English video-sentence/phrase pairs. To the best of our knowledge, this is the largest dataset for ISL. % processing. 
    \item \benchmarkName\ includes $3$ standard sign language processing  tasks: \textit{SignVideo2Text Translation, SignPose2Text translation, Text2Sign Translation, Sign/Gloss Recognition}. Two additional tasks of \textit{Sign Presence Detection} and \textit{Sign Semantic Similarity Prediction} are introduced. The two additional tasks are added to encourage representation learning and contextualized learning in Signed Languages. We develop baseline models and report results for each of the task. We release the data, tasks, and baseline models (\url{https://exploration-lab.github.io/iSign/}). %Moreover, upon paper acceptance, we plan to create a leaderboard (similar to SuperGLUE \cite{wang2019superglue} for NLP tasks) where researchers could submit and compare their model for each of the tasks. % and compare it with others.
    \item We conduct a detailed analysis of the ISL dataset and analyze how it differs from spoken languages. We provide linguistic insights into the functioning of ISL, covering various aspects like structural differences, the significance of non-manual markers, the use of space, the use of fingerspelling and co-reference, and role shifts. We hope that the detailed analysis will open up a new set of computational challenges from the linguistic perspective of ISL and further encourage various research directions. 
\end{itemize}

\section{Related Work} \label{sec:related}
% \vspace{-3mm}

%%%%%%%%%%%%%%%%%%%%%%%%%%%%%%%%%%%%%%%%%%%%%%
\begin{table*}[t!]
\small
\centering
\renewcommand{\arraystretch}{1}
\setlength\tabcolsep{2.5pt}
\begin{tabular}{lccccccc}
\toprule
\textbf{Datasets} & \textbf{Sign-Language} & \textbf{Words} & \textbf{Videos}      & \multicolumn{1}{c}{\textbf{Avg. Videos/ Word}} & \textbf{Signers}     & \textbf{Modalities}  & \textbf{Categories} \\
\midrule
Boston ASLLVD     & American               & 2742           & 9794                 & \multicolumn{1}{c}{3.6}                        & 6                    & RGB & -                   \\
DEVISIGN-L          & Chinese                & 2000           & 24000                &             12                                   & 8                    & RGB, depth           & -                   \\
DGS Kinnect       & German                 & 40             & 3000                 &                           75                     & 15                   & RGB, depth           & -                   \\
GSL               & Greek                  & 20             & 840                  &                               42                 & 6                    & RGB                  & -                   \\
LAS64             & Argentinian            & 64             & 3200                 &                            50                    & 10                   & RGB                  & -                   \\
LSE-sign          & Spanish                & 2400           & 2400                 &                               1                 & 2                    & RGB                  & -                   \\
Perdue RVL-SLLL   & American               & 39             & 546                  & \multicolumn{1}{c}{14}                         & 14                   & RGB & -                   \\
PSL Kinnect 30    & Polish                 & 30             & 300                  &                         10                  &  \multicolumn{1}{c}{-} & RGB, depth           & -                   \\
RWTH-BOSTON-50    & American               & 50             & 483                  &                         9.7                       & 3                    & RGB &  -                   \\
WLASL             & American               & 2000           & 21,083               & \multicolumn{1}{c}{10.5}                       & 119                  & RGB                  & -                   \\
\midrule
\citet{resource_indian_nandy2010recognition}             & Indian                 & 22             & 600                  &                                 27.3               & \multicolumn{1}{c}{-} & RGB                  & -                   \\
\citet{resource_indian_kishore2012video}           & Indian                 & 80             & 800                 &     10                                           & \multicolumn{1}{c}{-} & RGB                  & -                   \\
INCLUDE           & Indian                 & 263            & 4287                        & 16.3                                            & 7 & RGB                  & 15                  \\
ISL-CSLRT         & Indian                 & 186            & 700                  &                  3.8                              & 7                    & RGB                  & -                   \\
CISLR \citep{joshi-etal-2022-cislr}             & Indian                 & 4765           & 7050 & 1.5                                                & 71                   & RGB                  & 57                 \\
\bottomrule
\end{tabular}
  \caption{{
  % The proposed Indian-Sign Language Dataset comparison with other Sign-Language datasets.
  The Indian-Sign Language Dataset comparison with other Sign-Language datasets for Isolated Sign Language Recognition.
  }}
  \label{tab:islr-dataset-comparison}
  % \vspace{-2mm}
\end{table*}

\begin{table*}[t]
\centering
\small
\renewcommand{\arraystretch}{1}
\setlength\tabcolsep{6pt}
\begin{tabular}{lcccc}
\toprule
Dataset             & Language & Sentences & Vocab. (corresponding Text) & Hours \\ 
\midrule 
% Purdue RVL-SLLL 
\begin{tabular}[l]{@{}l@{}}Purdue RVL-SLLL \cite{RVL-SLLL-Martinez2002PurdueRA}\end{tabular}
& ASL      & 2.5k         & 104  & - \\
\begin{tabular}[l]{@{}l@{}}Boston 104 \cite{boston_104-dreuw07_interspeech}\end{tabular}
% Boston 104 \\\cite{boston_104-dreuw07_interspeech}          
& ASL      & 201          & 103   & - \\
% How2Sign            
\begin{tabular}[l]{@{}l@{}}How2Sign  \cite{How2Sign_Duarte_CVPR2021}\end{tabular}
& ASL      & 35k          & 16k   & 79 \\
% OpenASL             
\begin{tabular}[l]{@{}l@{}}OpenASL \cite{OpenASL}\end{tabular}
& ASL      & 98k          & 33k  & 288 \\ 

\begin{tabular}[l]{@{}l@{}}YouTube-ASL \cite{OpenASL}\end{tabular}
& ASL      & 610k          & 60k  & 984 \\ 

% AfriSign
\begin{tabular}[l]{@{}l@{}}AfriSign \cite{gueuwou2023afrisign}\end{tabular}
&    \begin{tabular}{@{}c@{}} KSL, ZSL, SASL \\ GSL, NSL, ZISL \end{tabular}  & 98k          & 20k  & - \\ 
% \begin{tabular}[l]{@{}l@{}}AfriSign \cite{gueuwou2023afrisign}\end{tabular}
% & GSL, NSL, ZISL, KSL, ZSL, SASL      & 98k          & 20k   \\ 

% BOBSL               
\begin{tabular}[l]{@{}l@{}}BOBSL \cite{BOBSL-Albanie2021bobsl}\end{tabular}
& BSL      & 993k         & 72k  & 1447 \\

% CSL Daily
\begin{tabular}[l]{@{}l@{}}CSL Daily  \cite{CSL_daily_dataset_Zhou2021ImprovingSL}\end{tabular}

& CSL      & 20.6k        & 2k   & 23 \\

% Phoenix-2014T
\begin{tabular}[l]{@{}l@{}}Phoenix-2014T  \cite{phoenix_dataset}\end{tabular}
& DGS      & 8.2k          & 3K   & 11 \\

% SWISSTXT-Weather
\begin{tabular}[l]{@{}l@{}}SWISSTXT-Weather \cite{Content4All-open-sign-language-datasets}\end{tabular}
& DSGS     & 811          & 1k   & - \\

% SWISSTXT-News       
\begin{tabular}[l]{@{}l@{}}SWISSTXT-News \cite{Content4All-open-sign-language-datasets}\end{tabular}
& DSGS     & 6k           & 10k  & -  \\

% KETI
\begin{tabular}[l]{@{}l@{}}KETI \cite{KETI_dataset_Ko2018NeuralSL}\end{tabular}
& KSL      & 14.6k        & 419   & 28 \\

% VRT-News            
\begin{tabular}[l]{@{}l@{}}VRT-News \cite{Content4All-open-sign-language-datasets}\end{tabular}
& VGT      & 7.1k         & 7k   & 100  \\
\midrule
%ISL-CSLRT    
 \begin{tabular}[l]{@{}l@{}}ISL-CSLRT 
 \cite{resource_indian_eelakkiya2021isl_ISL_CSLRT}
\end{tabular}
& ISL      & 100          &     -  & - \\
ISLTranslate \citep{joshi-etal-2023-isltranslate} & ISL      & 31k          & 11k   & 55 \\ 
\benchmarkName\ (ours) & ISL      &   \iSignCountShort        & 40k & 252    \\
\bottomrule
\end{tabular}
\caption{Comparison of continuous sign language translation datasets. Please refer to the App. Table \ref{app-tab:sign-languages-fullform} for details.
% \AJ{commented numbers to be updated to new numbers}
}% \AM{need to write abt Google's ASL dataset and also abt the large BSL dataset}. \RM{Readers are requested to refer to \ref{tab:sign_languages} for a detailed description DGS = Deutsche Gebärdensprache, CSL = Chinese Sign Language, BSL = British Sign Language, ASL = American Sign Language, GSL = Ghanaian Sign Language, NSL = Nigerian Sign Language, KSL = Kenyan Sign Language, ZSL = Zambian Sign Language, ZISL = Zimbabwean Sign Language, SASL = South African Sign Language, ISL =  Indian Sign Language, DSGS = Deutschschweizer Gebärdensprache(Swiss German Sign Language), VGT = Vlaamse Gebarentaa (Flemish Sign Language) }}
\label{tab:translation-comparison}
% \vspace{-6mm}
\end{table*}
%%%%%%%%%%%%%%%%%%%%%%%%%%%%%%%%%%%%%%%%%%%%%%

Recently, the research community has been actively interested in developing tools and techniques for processing sign languages. Since sign languages contain both visual, gestural, and language modalities, both the vision \cite{american_sign_language_WASL_li2020word} and natural language \cite{yin-etal-2021-including} research communities have developed techniques. Several tasks for sign language processing have been proposed, for example, sign language detection \cite{moryossef2020real}, identification \cite{monteiro2016detecting}, segmentation \cite{bull2020automatic}, recognition (gloss detection) \cite{imashev2020dataset,turkish_sign_language_sincan2020autsl}, generation \cite{saunders2020progressive,saunders2020everybody,xiao2020skeleton,survey}, and translation \cite{jiang-etal-2023-machine, muller-etal-2022-findings, Muller2022ConsiderationsFM, moryossef2021data,yin2020attention,yin2020better,camgoz2018neural,camgoz2020sign}.

\noindent\textbf{Isolated Sign Language Recognition/Gloss Recognition:}
Many benchmarks have been proposed for gloss recognition (\S\ref{sec:tasks}) in sign languages other than ISL \cite{resource_Mesch2012FromMT, resource_Fenlon2015BuildingBS, resource_LSE_Sign_gutierrez2016lse,resource_Martinez2002PurdueRA,resource_RWTH_BOSTON_zahedi05:dagm05,resource_efthimiou2007gslc_GSL,tavella-etal-2022-wlasl-lex}. There are very few datasets for ISL like \citet{resource_indian_rekha2011shape,resource_indian_nandy2010recognition,resource_indian_kishore2012video,selvaraj-etal-2022-openhands}, INCLUDE dataset \cite{resource_indian_INCLUDE}, ISL-CSLRT dataset \cite{resource_indian_eelakkiya2021isl_ISL_CSLRT}, CISLR \cite{joshi-etal-2022-cislr}, and ISLTranslate \cite{joshi-etal-2023-isltranslate}. Table \ref{tab:islr-dataset-comparison}) provides a comparison with other isolated sign language recognition datasets.  

\noindent\textbf{Sign Language Translation Datasets:} Various datasets \cite{yin-etal-2021-including} for sign language translation have been proposed in recent years for different sign languages, e.g., ASL \cite{RVL-SLLL-Martinez2002PurdueRA,boston_104-dreuw07_interspeech, uthus2023youtubeasl}, Chinese sign language \cite{CSL_daily_dataset_Zhou2021ImprovingSL}, Korean sign language \cite{KETI_dataset_Ko2018NeuralSL}, Swiss German Sign Language - Deutschschweizer Gebardensprache (DSGS) and Flemish Sign Language - Vlaamse Gebarentaal (VGT) \cite{Content4All-open-sign-language-datasets}. Table \ref{tab:translation-comparison} provides a comparison with datasets for other sign languages. 

\noindent\textbf{Sign Language Generation:}
The field of sign language production predominantly revolves around the generation of hand movements, with the majority of existing approaches leveraging GAN architectures and sometimes in combination with transformer architecture \cite{DBLP:journals/ijcv/StollCHB20,signingatscale,saunders2020progressivetransformers,9093516,DBLP:journals/corr/abs-2008-12405}.

\section{\benchmarkName\ Benchmark} \label{sec:benchmark}
% \vspace{-3mm}

%In this section, we describe the process for creating \benchmarkName\ benchmark. 
%%%%%%%%%%%%%%%%%%%%%%%%%%%%%%
\begin{table*}[t]
\centering
\small
\renewcommand{\arraystretch}{1.1}
\setlength\tabcolsep{1pt}
\begin{tabular}{ll}
\toprule
\;\;\;\;\;\;\;\;\;\;\;\;\;\;\;\;\;\;\;\;\textbf{Dataset Translations}                                   & \;\;\;\;\;\;\;\;\;\;\;\;\;\;\;\textbf{ISL-Signer Translations (references)}
\\ \midrule
% Birbal started smiling. When it was his turn, he went near the line. & Birbal started smiling. He \textcolor{blue}{turned towards the drawn line.}              \\
Where are you going to, man? I \textcolor{red}{said}.                    & Where are you going to, man? I asked.                        \\
Where are you going this \textcolor{red}{fine} day? I \textcolor{red}{said to} the puppy.                                          &Where are you going  \textcolor{blue}{on} this \textcolor{blue}{special} day? I \textcolor{blue}{asked} the puppy.                                           \\
\textcolor{red}{and} name your dog.                                                 & name your dog.                                                   \\
% He didn't touch the first line. Everyone in the court.               & He didn't touch the first line. Everyone in the court saw what he drew \\
\textcolor{red}{You might begin}. I am a little brown dog.                                          & \textcolor{blue}{What do you begin?} I am a little brown dog.                                                          \\
Not I.              & Not\textcolor{blue}{,} \textcolor{blue}{I did not come}.                          \\
I \textcolor{red}{said} to the horse as he went by.           & I \textcolor{blue}{asked }the horse as he went by.                                \\
\textcolor{red}{as he went by, up in the hills }                                           &   \textcolor{blue}{I asked the puppy as he went by.}                                                \\
 \textcolor{red}{Autobiography}                                       &  \textcolor{blue}{to tell your life story in your own language.}                                  \\
Page 117.  \textcolor{red}{It was impossible for me to...              }& Page 117.  \textcolor{blue}{The two fences                }                  \\
climb because \textcolor{red}{every step} was 6 feet high.                                    & \textcolor{blue}{were impossible for me to} climb because \textcolor{blue}{they were} 6 ft high.                              \\
\textcolor{red}{60 feet} above the ground.                                & \textcolor{blue}{I was high} above the ground \textcolor{blue}{at}                \\
\textcolor{red}{and} blew my hair aside to get a better view of my face.                                             & \textcolor{blue}{He} blew my hair aside to get a better view of my face.                                               \\
Each minister looked at the line and was puzzled.                    & Each minister looked at the line and was puzzled.                      \\
No one could think of any way to make it longer.                     & No one could think of any way to make it longer.                       \\
\textcolor{red}{I} turned back to join \textcolor{red}{the} crew.                                & \textcolor{blue}{Gulliver} turned back to join \textcolor{blue}{his} crew.                                              \\
\textcolor{red}{Dinner was} brought \textcolor{red}{for the} farmer in a dish.                                     & \textcolor{blue}{A} farmer brought food in a dish                                    \\
\textcolor{red}{Land with} no vegetation.                                                       & \textcolor{blue}{meaning is} no vegetation.                                                              \\
and some had wells to supply water.                                  & and some had wells to supply water.                                    \\
\textcolor{red}{My} wife and my children.                           & wife and my children. \textcolor{blue}{had covered drains.           }                    \\
% Notice how carefully these were laid out in straight lines.          & Notice how carefully these were laid out in straight lines.           \\ 
\bottomrule
\end{tabular}
\caption{The Table shows a sample of English translations present in the created dataset compared to sentences translated by ISL Signer. \textcolor{blue}{Blue} and \textcolor{red}{Red} colored text highlight the difference between semi-automatically generated English sentences and gold sentences generated by the ISL instructor.}% \AM{change the examples and these should be different from ISLTranslate}}
\label{tab:islrtc-reference-sample-sentences}
% \vspace{-3mm}
\end{table*}

\begin{table*}[t]
\centering
\small
\renewcommand{\arraystretch}{1.1}
\setlength\tabcolsep{2pt}
\begin{tabular}{ccccccccccc}
\toprule
\textbf{Metric} & \textbf{BLEU-1} & \textbf{BLEU-2} & \textbf{BLEU-3} & \textbf{BLEU-4} & \textbf{METEOR} & \textbf{WER} & \textbf{ROUGE-L}  & \textbf{ROUGE-1} & \textbf{ROUGE-2} & \textbf{ROUGE-L-SUM}\\
\midrule
\textbf{Score} & 76.3 & 73.42 & 71.2 & 69.3 & 73.83 & 33.83 & 81.9 & 82.2 & 64.4 & 81.8 \\

\bottomrule
\end{tabular}
\caption{The Table shows the Translation scores for a sample of 593 sentence pairs from the created dataset when compared to references translated by ISL Signer.}% \AM{these numbers need to be revised for new annotations}} \AJ{BLEU scale does not match with table 5, should be consistent, use either [0-1] or [0-100] for BLEU scores}
\label{tab:islrtc-reference-translations-metrics}
% \vspace{-5mm}
\end{table*}
%%%%%%%%%%%%%%%%%%%%%%%%%%%%%%
% \noindent\textbf{Dataset Creation:}
% To create the translation dataset with ISL-English video-sentence/phrase pairs, we primarily use three publicly available resources: ISLRTC videos,\footnote{\url{https://www.youtube.com/@islrtcnewdelhi4069/playlists}} ISH News (News channel in ISL),\footnote{\url{https://www.youtube.com/channel/UC99w_Bzj8ikOz8Gpv0prbNg}} and phrases from DEF (Deaf Enabled Foundation).\footnote{\url{https://www.youtube.com/channel/UCM7U7CyJGRIbu4qsmSh3UZg}}. These YouTube channels provide permission to scrape videos and use them for research. \TB{The sign language videos come with CC BY-ND 4.0 license (non-commercial research only).} The scraped videos contain a signer communicating a sentence to the spectators. Note that we ignore all the video contents where two signers communicate with each other, as in sign language, a two-signer communication might use different sets of reference points for communication. Moreover, in such videos, the orientation of the signer might change completely, resulting in more variation from the visual perspective due to different camera angle placement. We speculate a keypoint detection framework might be helpful here as the 3D key points can be normalized to keep the same viewing angle; however, the problem of different gesture/sign usage for change in reference point will remain. The videos are split and pre-processed to obtain video-sentence/phrase level pairs. The exact splitting, the pre-processing process and data statistics are described in App. \ref{app:dataset-details}. Fig. \ref{fig:iSign_thumbnail} shows an example from \benchmarkName. 

\noindent\textbf{Dataset Creation:}
%For creating \benchmarkName, we primarily use three publicly available and authentic resources on YouTube: ISLRTC videos\footnote{\url{https://www.youtube.com/@islrtcnewdelhi4069/playlists}} (a Government of Indian initiative), ISH News (News channel in ISL),\footnote{\url{https://www.youtube.com/channel/UC99w_Bzj8ikOz8Gpv0prbNg}} and phrases from DEF (Deaf Enabled Foundation, a non-profit working for the DHH community).\footnote{\url{https://www.youtube.com/channel/UCM7U7CyJGRIbu4qsmSh3UZg}}  h. The sign language videos come with CC BY-ND 4.0 license (non-commercial research only)
For creating \benchmarkName, we primarily use three publicly available and authentic resources on YouTube: ISLRTC videos\footnote{\url{http://tinyurl.com/mr3v4ead}} (a Government of Indian initiative), ISH News (News channel in ISL),\footnote{\url{http://tinyurl.com/4a9xe6rk}} and phrases from DEF (Deaf Enabled Foundation, a non-profit working for the DHH community).\footnote{\url{http://tinyurl.com/3rzw7xff}} 
These YouTube channels provide permissions to scrape videos and use them for research. Each of these videos contains a single signer communicating information (educational content or news about current affairs) in ISL and a corresponding transcript in English. The videos are pre-processed and split to obtain video-sentence level (or phrase level) pairs. Fig. \ref{fig:iSign_thumbnail} shows an example from \benchmarkName. The exact splitting, the pre-processing process, and data statistics are described in the App. \ref{app:dataset-details}. Since the YouTube channels keep getting populated with new content, we will continue to grow \benchmarkName\ with more video-sentence pairs. Note that in the current version of the benchmark, we ignore all the videos where two signers communicate with each other, as in sign language, a two-signer communication might use different sets of reference points for communication. Moreover, in such videos, the orientation of the signer might change completely, resulting in more variation from the visual perspective due to different camera angle placement. We speculate a key-point detection framework might be helpful here as the 3D key points can be normalized to keep the same viewing angle; however, the problem of different gesture/sign usage for change in reference point will remain. We leave the exploration of including multiple signer videos for future work.

\noindent\textbf{Comparison with Existing Datasets:} Table \ref{tab:islr-dataset-comparison} shows word-level datasets and Table \ref{tab:translation-comparison} compares \benchmarkName\ with other video-based sign language datasets. For ISL, there are two existing publicly available datasets: ISLTranslate \citep{joshi-etal-2023-isltranslate} (having $31k$ video-sentence pairs) and CISLR \citep{joshi-etal-2022-cislr} (having 7k video-word pairs) for Isolated Sign Language Recognition. These two datasets were the largest among the previous ISL datasets. We also include these ISL datasets in \benchmarkName, after due permissions from the authors. It results in \iSignCount video-English sentence/phrase/words pairs in \benchmarkName. In the past, other works like \citet{uthus2023youtubeasl, OpenASL} followed a similar strategy of using YouTube videos with captions for generating translation datasets.

\noindent\textbf{Validation:}
To verify the reliability of the video-sentence/phrase ISL-English pairs present in the dataset, we took the help of three certified ISL signers. The signers worked with us on a pro-bono basis (details in App. \ref{app:annotation}). Due to the limited availability of certified ISL signers, we could only use a small randomly selected sign-text pairs sample ($593$ pairs) for human translation and validation. We asked ISL instructors to translate the videos. Each video is provided with one reference translation by the signers. Table \ref{tab:islrtc-reference-sample-sentences} shows a sample of sentences created by the ISL instructor. To quantitatively estimate the reliability of the translations in the dataset, we compare the English translation text present in the dataset with the ones provided by the ISL instructors. Table \ref{tab:islrtc-reference-translations-metrics} shows the translation scores for $593$ sentences in the created dataset. Overall, the BLEU-4 score is $69.3$ (indicative of high reliability), ROUGE-L \cite{lin-2004-rouge} is $81.9$, and WER (Word Error Rate) is $33.83$. To provide a reference for comparison, for text-to-text translations, the BLEU score of human translations ranges from 30-50 as reported by \citet{papineni-etal-2002-bleu}. Ideally, it would be better to have multiple reference translations available for the same signed sentence in a video; however, the high annotation effort and the lower availability of certified ISL signers make it a challenging task.

\section{\benchmarkName\ Tasks} \label{sec:tasks}
% \vspace{-2mm}

We propose various tasks in \benchmarkName\  to evaluate and compare different models developed for ISL processing. The tasks are described below.

\noindent\textbf{\noindent{Task 1) ISL-to-English Translation:}} It is a standard task of translating a sign (source) language to a spoken (target; in text form) language. As done in previous work, we use standard neural machine translation metrics to benchmark the baseline models on this task, including BLEU, METEOR, and ROUGE-L. Input (modality) to the translation system is a video in the form of a sequence of RGB images or pose-based features. There is a significant difference in performance based on the input modality. Hence, we add two subtasks under this task to facilitate the development and comparison of various approaches: \textbf{ISLVideo-to-English Translation} and  \textbf{ISLPose-to-English Translation}. The former uses image-based features as input \cite{camgoz2020sign, OpenASL,chen2022two-stream,cheng2023cico}, and the latter uses pose-based features in the form of body key points \cite{uthus2023youtubeasl,selvaraj-etal-2022-openhands}. Image-based approaches have shown promising results for sign language translation tasks. However, image-based architectures are compute-heavy and require more time for inference. Moreover, regarding large-scale application perspective, including images may result in signer-based biases creeping into the model learning. In contrast, extracting body pose features is fast and easy on edge devices; hence, pose-based approaches are more practical. 

\noindent\textbf{\noindent{Task 2) English-to-ISLPose Generation:}} The goal of this task is to transform textual input into a sequence of body poses that correspond to the sign language representation of the input sentence \cite{saunders2020progressivetransformers}. Hence, this task aims to generate a sign language video. The generated translations are evaluated using the Dynamic Time Warping (DTW) metric \citep{muller2007dynamic}. The DTW algorithm measures the alignment between the generated pose sequence and the ground truth, allowing us to assess how well the generated poses match the expected sign language representation. 

\noindent\textbf{\noindent{Task 3) Word/Gloss Recognition (Isolated Sign Recognition):}} Given a video of a signer (performing gestures and actions), the task is to predict the corresponding gloss label (word). We follow an existing work, CISLR \cite{joshi-etal-2022-cislr}, which contains a low number of average videos per word (1.5 videos per word), and formulate the gloss recognition task as a one-shot learning task. Overall, the CISLR task contains $4765$ sign video samples, which act as prototypes, and the task is to classify the remaining $2285$ videos into one of the $4765$ categories. We consider the standard metric of Top-1, Top-5, and Top-10 classification accuracy scores to evaluate this task.

\noindent\textbf{\noindent{Task 4) Word Presence Prediction:}} To capture the quality of sign representations learned by algorithms, we define a new task of Word Presence Prediction. Given a pair of a word (as a query) and a sentence (as a candidate) as two signed ISL videos, the task is to predict if the query word is present (used) in the signed sentence. For similarity comparison, we consider the cosine similarity of the representations obtained for the pair of ISL videos.
% \textcolor{orange}
{We evaluate the performance over this task using the standard classification accuracy, i.e., if the learned representations are able to predict the word presence given a query and a candidate pair. Since this task can also be treated as a retrieval task, we also consider Top Rank (Avg.) by ranking the entire pool of candidates for a particular query. Note that the better the rank, the better the representations learned by the model. (more details in the App. \ref{app:metrics})}
% We evaluate the performance of this task using standard classification accuracy and cosine similarity. \AJ{have to add the details based on the results and reported metrics}
% \AJ{refer Appendix for the data creation process, as mentioned in the review}

\noindent\textbf{\noindent{Task 5) Semantic Similarity Prediction:}} Given a pair of a word (as a query) and a sentence (as a candidate) as two signed ISL videos, we propose a new task of predicting the semantic similarity of videos. We select the ISL description videos corresponding to an ISL word as the candidate. For example, a sample for this task will contain an ISL video for the word “revision,” the corresponding ISL sentence will describe the same word, “a change that is made to something, or the process of doing this.” 
For similarity comparison, we consider the cosine similarity of the representations obtained for the pair of ISL videos. 
% \textcolor{green}
{To evaluate the representations learned by the models, we determine if the correct match is present within the top 5\% of the cosine similarity scores of the query-candidate pairs. Additionally, we consider the rank metric, indicating the position of the true match within this ranked list. More details about the evaluation metrics can be found in the App. \ref{app:metrics}}  
% \AJ{have to add the details based on the results and reported metrics}
% \AJ{refer Appendix for the data creation process, as mentioned in the review}

\noindent Overall, Task 1, Task 2, and Task 3 correspond to standard tasks proposed for processing other sign languages as well; nevertheless, given the scale of data, the tasks are new for ISL. Task 4 and Task 5 are newly introduced in this paper to promote representation learning in ISL. Table \ref{tab:data-details} summarizes the number of samples present for each task in the \benchmarkName\ benchmark.
% \AJ{should we explain the dataset splits at the start of this paragraph}

\begin{table}[t]
\centering
\small
\renewcommand{\arraystretch}{1.1}
\setlength\tabcolsep{2pt}
\begin{tabular}{clr}
\toprule
\textbf{Task ID} & \textbf{Task Name} & \textbf{\# Samples} 
% & \textbf{Number of Dev Samples} & \textbf{Number of Test Samples} 
\\
\midrule
1. & ISL-to-English Translation & \iSignCount
% & & 
\\
2. & English-to-ISL Pose Generation & \iSignCount
% & & 
\\
3. & Word/Gloss Recognition & $7,050$
% & & 
\\
4. & Word Presence Prediction & $1,523$
% & & 
\\
5. & Semantic Similarity Prediction & $593$
% & & 
\\
\bottomrule
\end{tabular}
\caption{Number of samples present in the \benchmarkName\ benchmark for various tasks. 
Note that tasks 4 and 5 are only used for validating the quality of the learned representation and have no trainset.
}
\label{tab:data-details}
\end{table}

\vspace{-1mm}
\section{Models, Experiments and Results} \label{sec:models}
\vspace{-2mm}

\begin{table*}[t]
\centering
\small
\renewcommand{\arraystretch}{0.9}
\setlength\tabcolsep{0.1pt}
\begin{tabular}{ccccccccc}
\toprule
%\toprule
\multicolumn{9}{c}{\textbf{Neural Machine Translation}}                                                                                                                                                                                                              \\ 
\midrule
\textbf{Task}                      & \textbf{Source Language} & \textbf{Target Language}        
% & \textbf{\# Test} 
& \textbf{Model} & 
% \textbf{BLEU-1}           & \textbf{BLEU-2}           & \textbf{BLEU-3}            & 
\textbf{BLEU-4} & \textbf{ROUGE-L} \\
\midrule
SignVideo-to-Text         &             ISL             &         English                        
% & 8078
&   \citet{camgoz2020sign} &      
% 13.36      &  4.54                           &        1.56                   &       
0.56                     &        \textbf{19.58}                          \\
SignVideo-to-Text         &             ISL             &         English                        
% & 8078
&  T5(small)+I3D(2000) &      
% 13.36      &  4.54                           &        1.56                   &       
0.24                     &        11.41                          \\
SignPose-to-Text                   &     ISL                   &       English                         
% & 8078                      
&    \citet{camgoz2020sign} &      
% 11.20      &                  3.78          &              1.53             &    
0.77                        &          9.52                         \\
SignPose-to-Text                   &     ISL                   &       English                         
% & 8078                      
&    T5(large)+Mediapipe(75) &      
% 11.20      &                  3.78          &              1.53             &    
0.09                        &         19.11                       \\
SignPose-to-Text                   &     ISL                   &       English                         
% & 8078                      
&    T5(small)+Mediapipe(75) &      
% 11.20      &                  3.78          &              1.53             &    
0.8                       &          16.46                       \\
SignPose-to-Text                   &     ISL                   &       English                         
% & 8078                      
&    T5(base)+Mediapipe(75) &      
% 11.20      &                  3.78          &              1.53             &    
\textbf{1.47}                       &          16.67                       \\
SignPose-to-Text                   &     ISL                   &       English                         
% & 8078    
&    SLT+Mediapipe(75) &      
% 11.20      &                  3.78          &              1.53             &    
0.36                       &          7.60                       \\
\toprule
\toprule
\multicolumn{9}{c}{\textbf{Sign Language Generation}}                                                                                                                                                                                                 \\ \midrule
 \textbf{Task}                      & \textbf{Source Language} & \textbf{Target Language}        & 
 % \textbf{\# Test} & 
 \textbf{Model} & \textbf{DTW}           &            &               &\\ \midrule

 Text-to-SignPose                   &        English                  &     ISL                         
 % & 8078
 &       \citet{saunders2020progressive}         &     22.69                      &                           &                                              \\
\toprule
\toprule
\multicolumn{9}{c}{\textbf{Word Level Translation/Gloss Prediction}}                                                                                                                                                                                                 \\ \midrule
 \textbf{Task}                      & \textbf{Source Language} & \textbf{Target Language}        & 
 % \textbf{\# Test} & 
 \textbf{Model} & \textbf{Acc. (Top-1\%)} & \textbf{Acc. (Top-5\%)} & 
 % \textbf{Acc. (Top-10)} &
 \\ \midrule
ISLR & ISL                      & English/Gloss                   
% & 2285
&  \citet{joshi-etal-2022-cislr}              & 16.81                           &   20.04        
% &  22.58
\\

\toprule
\toprule
\multicolumn{9}{c}{\textbf{Sign Representation Learning}}                                                                                                                                                                                                                 \\ \midrule
\textbf{Task}                      & \textbf{Query}           & \textbf{Candidate}              & 
% \textbf{\# Test} & 
\textbf{Model} & \textbf{Top 5\% Acc.}         & 
% \textbf{Precision}        & \textbf{Recall} 
% & \textbf{F1}
% & 
\textbf{Rank (Avg.)}                \\ \midrule
Word Presence           & word           & example sentence &  T5-base + I3D                          &  $52$                &     $193/1523$                      &                           &                            &                                   \\ 
\midrule
Semantic Similarity           & word           & description sentence &  T5-base + I3D                          & $67$               & 44/593                           &                           &                            &                                    \\ 
% \midrule

% \multicolumn{2}{c}{\textbf{Semantic Similarity}}   

%                             \\    \midrule
% \textbf{Semantic Similarity}       &  \textbf{Cosine Similarity (Avg.)}                \\ \midrule

% Transformer Architecture     &                  63.67\%         &            \\                  
% \toprule
% \toprule
% \multicolumn{10}{c}{\textbf{Sign Retrieval}}                                                            \\
\bottomrule
%\bottomrule
\end{tabular}
\caption{Results of various baseline models on proposed tasks.}
\label{tab:results_main}
% \vspace{-5mm}
\end{table*}

%\subsection{Baseline Models}
% \vspace{-4mm}
\noindent\textbf{Baseline Models:}
We experimented with various models for the proposed task as described next (model training details provided in App. \ref{app:models-details}). 

\noindent1) \underline{ISL-to-English Translation:}
For this task, we follow \citet{camgoz2020sign} and validate the performance for both the sub-tasks. For ISLVideo-to-English Translation, we use spatial embeddings extracted from pre-trained CNNs as input for our model. For ISLPose-to-English Translation, we follow \citet{saunders2020progressivetransformers} and create a sequence of poses to act as input to the SLT (Sign Language Transformer) model \cite{camgoz2020sign}. For pose key points, we use the Mediapipe pose estimation pipeline \cite{MediaPipe}.

\noindent2) \underline{English-to-ISLPose Generation:}
We utilize a transformer-based architecture as introduced in \citet{saunders2020progressivetransformers} to generate body key points corresponding to the textual input. 

\noindent3) \underline{Word/Gloss Recognition:} For this task, we follow CISLR \cite{joshi-etal-2022-cislr} as the baseline. CISLR uses the state-of-the-art model (Inception3D (I3D) \citep{carreira2017quo_I3D}) on the WLASL dataset \citep{WLASL} and trains it on 2000 classes. Further, the penultimate layer of the trained model is used to generate features corresponding to the prototype videos in the dataset, and each test sample video is assigned the gloss corresponding to the nearest prototype using cosine similarity between the obtained features.  

\noindent4) \underline{Word Presence Prediction:}
As the motivation of this task is to validate the representations learned by a neural architecture, we consider a pre-trained I3D, trained on the Human Kinetics Dataset \citep{kay2017kinetics}, as a baseline and report the findings.

\noindent5) \underline{Semantic Similarity Prediction:}
We use average cosine similarity scores between the sign representations of the word and their corresponding description to measure the similarity. The representations are obtained using a pre-trained I3D network \cite{carreira2017quo_I3D}.

\noindent\textbf{Results:}
Table \ref{tab:results_main} shows the baseline results for all the tasks in \benchmarkName. The BLEU scores for Sign-to-Text translation (Task 1) are a bit low,  
though we follow the previous baselines \cite{camgoz2020sign} that perform well (with a BLEU score of $\sim 20$) on the RWTH Phoenix 2014T dataset \cite{phoenix_dataset}. We speculate a high gap between the two datasets to be the primary reason for the observed difference. For Task 2, we obtained a DTW score of 22.69, pointing towards high variation in generated and ground truth poses. For Task 3, we found the I3D features used by \citet{joshi-etal-2022-cislr} to be performing better than the representations extracted via the models trained for Task 1. For Task 4 and 5, we compute the Top 5\% accuracy by performing a one vs all prediction, i.e., a query's feature representation will have the highest similarity with the corresponding candidate's feature representation. We also report the average rank assigned via similarity corresponding to all the candidates (i.e., the lower the rank, the better the learned representations). For Task 4, we found the T5base + I3D features getting Top-5\% Acc. of $52\%$ and average rank as $193$ out of $1523$ candidates (also check App. Table \ref{app-tab:results-task3-4-5}). Overall, all the baseline performance points towards a huge scope of future developments in ISL processing (also see Limitations section). 

\noindent\textbf{Poor Performance of Existing Baseline Models:} The current baseline models for several tasks show poor performance, pointing toward developing more sign-language-specific neural architectures in the future. The machine translation baselines, which show SOTA performance on datasets like Phoenix-2014T and CSL-Daily, do not perform well for the created dataset. Most existing approaches for sign language translation \cite{phoenix_dataset, info13050220, De_Coster_2021_AT4SSL, chen2022two-stream} depend on intermediate gloss labels for translations. As glosses are aligned to video segments, they provide fine one-to-one mapping that facilitates supervised learning in learning effective video representations. Previous work has reported a drop of about $10.0$ in BLEU-4 scores without gloss labels \cite{phoenix_dataset}. However, considering the annotation cost of gloss-level annotations, it becomes imperative to consider gloss-free sign language translation approaches. Moreover, gloss mapping in continuous sign language might remove the grammatical aspects of sign language. The presence of gloss labels for sign sentences in a dataset helps translation systems work at a granular level of sign translation. However, generating gloss annotations for a signed sentence is an additional challenge due to the scarcity of certified signers. The large number of samples in the created dataset makes the gloss-level annotation infeasible. There are a few recent works on Sign language translation \citet{Voskou2021StochasticTN, yin-read-2020-better}, which try to remove the requirement for a glossing sequence for training and propose a transformer-based architecture for end-to-end translations. Moreover, the noteworthy point is that the dataset size of \benchmarkName\  differs from Phoenix-2014T and CSL-Daily by a significant margin (\iSignCountShort\ vs. 8.2k, 20.6k), making \benchmarkName\  more diverse with high variance in gestures/signs, which might also be one of the reasons for poor performance. Furthermore, in a recent work, \citet{muller-etal-2022-findings,muller-etal-2023-considerations} report the current performance of automatic translation systems to be very low compared to the text-based translation systems for spoken languages.

\noindent\textbf{Evaluation Metric for Sign Generation:} In the current version of  \benchmarkName, we use the Dynamic Time Warping (DTW) algorithm to evaluate the quality of generated sign pose key points. The DTW algorithm has limitations in dealing with significant variations in motion between the generated and ground truth sequences. Moreover, the DTW may not be suitable for measuring the quality of generated sign language. While designing the evaluation metrics, it is essential to consider sign language's linguistic aspects and ensure they correlate strongly with human evaluation (more details in App. \ref{app:metrics}).

\section{ISL Linguistics and Computational Challenges} \label{sec:linguistics}
% \vspace{-3mm}
%Sign language functioning differs from spoken languages by a significant margin. In this section, we highlight some ISL-specific features that might be important to consider while building systems for machine translation in ISL. We create this list of insights after having discussions with one of the authors (Dr. Andesha Mangla) who is a Professor of Sign Linguistics at ISLRTC. The selected list of these unique features will motivate the development of dedicated sign-language-specific neural architecture and facilitate understanding of ISL in the NLU research community.

Sign language functioning differs from spoken languages by a significant margin. In this section, we highlight some ISL-specific features that might be helpful in the development of dedicated sign-language-specific neural architecture and facilitate understanding of ISL in the NLP research community. 
% We created this list of insights after discussing it with a professor of Indian sign language linguistics.  \AJ{should we mention the name now in the camera-ready version?}
We created this list of insights after discussing it with a Professor of Indian Sign Language linguistics, Dr. Andesha Mangla, who is also one of the co-authors of this paper.

\noindent\textbf{Structural Differences:} ISL is a form of visual language that consists of signs, gestures, fingerspelling, and facial expressions going in parallel to communicate a sentence, making it quite different from spoken language in the structural form \cite{sinha2017indian}. At a rudimentary level, the building blocks of sign and spoken languages differ. In spoken languages, a combination of sounds results in the formation of words, while in sign languages, a mixture of manual and non-manual parameters forms words. Moreover, the usage of visual-spatial and manual modality in sign languages allows the production of various concepts in parallel. For example, using physical space for multiple purposes, using head, eyes, and body to represent different entities, actions, etc., and using non-manual expressions for various concepts. In terms of linguistics, iconicity plays a more significant role in the production and perception of sentences when compared to spoken languages \cite{Zeshan-9789027298522, sinha2017indian,brentari_2019}. 

\noindent\textbf{Significance of Non-Manual Markers:}  In ISL, non-manual markers like facial expressions, body language, etc., play a vital role in giving semantics to the produced sentences, both at lexical as well as grammatical levels \cite{sinha2017indian}. For example, the word ``HAPPY'' is signed with a smiling face, whereas the word ``SAD'' is signed with a sad facial expression. The order of non-manual markers goes in parallel with the manual markers, giving the sentence meaning. For example, the same sentence signed with a forward head tilt and wide-open eyes will transform the statement sentence into a yes-no question. Moreover, non-manual markers are also used in the production of complex sentences, such as conditional sentences. The various use cases and parallel nature of non-manual markers in sign language production make it more challenging for a sequential language-based model to adapt to ISL.

\noindent\textbf{Use of Space in ISL:} Physical signing space is crucial in making production and communication more efficient in ISL \cite{sinha2017indian}. Physical singing space provides a medium for assigning various reference points required in a specific sentence, like referring to designated locations for people, places, or any topic/subject. While communicating a sentence, the referents in a narration are assigned various locations in the signing space, which are then referred to in the conversation using the pointing sign toward the same space. Space provides a medium for references and grounds the language to actual space. For example, the word ``AEROPLANE'' will be signed in the upper portion of the signing space, whereas the word ``CAR'' will be signed in the lower portion of the signing space. As the references are created specific to sentences for each conversation, the linguistic structure of the language becomes more complex, and the conversation becomes challenging to separate into independent sentences.

\noindent\textbf{Fingerspelling and Co-reference:} In ISL, names for characters, places, etc., are produced in various ways. For introducing a new name in the conversation, the name is fingerspelled and simultaneously assigned a short sign consisting of the initials, which refers to the same name in future sentences \cite{sinha2017indian}. For example, a girl character named ``Neha'' is introduced in the conversation by signing ``FEMALE+CHILD (= girl) NAME N-E-H-A (fingerspelled) SIGN SHORT FEMALE+N'' at the start of the conversation. Later on, the name Neha is co-referenced with the assigned short sign ``FEMALE+N.'' Note that assigning a short sign is not unique and varies. For example, the same name, Neha, can be given a short sign using visual features and physical characteristics. For example, if there is a picture available of the child Neha, in which she is wearing a pair of spectacles, the sign introduction can look like ``FEMALE+CHILD (= girl) NAME N-E-H-A (fingerspelled) SIGN SHORT FEMALE+SPECS,'' where ``FEMALE+SPECS'' becomes the assigned short sign. Moreover, other variations can exist, combining the first and second examples to create a short sign ``N+SPECS,'' N coming from the name, and SPECS for the visual feature of spectacles. For co-referencing, ISL also makes use of signing space \cite{sinha2017indian}. For example, a character, place, etc., is given a location in the signing space during the introduction, and co-references are then made by pointing toward that location in space. For introducing new concepts, things, actions, etc., if there is no available sign, the signer describes the concept by fingerspelling it. 

\noindent\textbf{Role Shifts in ISL:} The visual modality provides signed languages with multiple ways of speaking the same sentence. This is similar to audio in spoken languages, where a speaker makes use of voice modulations to enact another person's role. In ISL, Signers use role shifts to indicate different entities \cite{sinha2017indian}. In role shift, the signer takes on the roles of the different participants in a narrative and enacts their roles \cite{lillo201217}. The respective roles are indicated by head and shoulder position, eye gaze, and non-manual expressions. This unique property of Role Shifts makes translation more challenging. As in written languages, the role shifts are generally indicated by a pretext like ``X said in a soft tone,'' followed by the respective dialogue. 

\noindent\textbf{Demographics and Dialects:} ISL, being used in a diverse country like India, incorporates numerous variations due to regional, cultural, and geographic diversities. Some of the anecdotal evidence indicates that eastern regions like West Bengal and southern regions like Tamil Nadu and Kerala have a higher degree of variation when compared with the Delhi (northern region) dialect of ISL \cite{jepson1991urban,zeshan2003indo,johnson2008assessment,Indian-Sign-Language-book}. Apart from the location/region of usage, socio-economic factors like age, gender, education, etc., have been shown to affect the dialects of ISL, leading to more variations. Anecdotal evidence indicates that around $75\%$ vocabulary is found to be similar across India, and the variation in the remaining $25\%$ is majorly found in categories like numbers, colors, months, weekdays, kinship terms, food, etc. \cite{jepson1991urban,zeshan2003indo,johnson2008assessment,Indian-Sign-Language-book}. 
% We provide additional linguistic aspects in App. \ref{app:linguistics}.

\noindent\textbf{Extended Usage of Verbs and Nouns:} 
% \textcolor{orange}
{Apart from the co-referencing feature discussed in the main paper, the usage of verbs and nouns is expressed in different ways depending on the actual physical characteristics of the objects. For example, ISL has a generic sign for OPEN. However, when talking about specific objects, such as opening a drawer, opening (starting) a computer, opening one’s eyes, etc., the sign gets modified depending on the object.}

\noindent\textbf{Use of Classifiers:} 
% \textcolor{orange}
{In Indian Sign Language, classifiers describe locations and movements. For example, a classifier for vehicles can be used to depict multiple things like their location, direction of motion, manner of movement, etc. The entity the classifier refers to depends on the context. For example, if the classifier is used after signing CAR, it will refer to CAR, but if it is signed after BUS, it will refer to BUS. The use of classifiers enables sign languages to describe visual details, manner of movement, etc., in greater detail than can be done in spoken languages. }

\noindent\textbf{Simultaneous Articulation:} 
% \textcolor{orange}
{The use of multiple articulators in parallel allows a signer to represent different entities simultaneously. For example, the left hand can depict a bridge, and the right hand can show a car moving under the bridge. }

We believe the working and insights of sign-language linguistics would promote the incorporation of domain knowledge into computational methods for sign-language processing. 

\section{Conclusion and Future Directions} \label{sec:conclusion}
% \vspace{-3mm}

We propose \benchmarkName, a benchmark for Indian Sign Language Processing, to bridge the gap between developments in spoken languages and signed languages and provide a standardization medium for accelerating the improvement. We release one of the largest available datasets for ISL with \iSignCountShort video-sentence/phrase pair sentences to facilitate research. We believe the released dataset will not only help improve translation models but also open up various ways for advancing natural language modeling techniques like contextualized representation learning, mask language modeling, capturing semantic similarities, etc., and encourage research in the NLP community. As a part of the benchmark, we incorporate various sign language tasks. In the future, we plan to grow the dataset by adding more samples. We also plan to add more ISL-based tasks (e.g., Sign Sentence Retrieval) to the benchmark. Additionally, we provide some linguistic insights into the functioning of ISL and discuss the open challenges in sign language processing. Accordingly, we plan to incorporate linguistic priors into the models. %In the future, we plan to keep a public leaderboard for easier replication and comparison of the proposed methods. % for ISL. 

\section*{Acknowledgements} \label{sec:acknowledgements}

We would like to thank anonymous reviewers for their insightful comments. We would like to thank the Indian Sign Language Research And Training Center (ILSRTC) team for helping us validate the quality of the curated translation dataset. We would also like to extend our immense gratitude towards the ISLRTC members for providing us full support in sharing the ISL content creation process.
Finally, we thank the ISL content creators on YouTube (ISH-TV and DEF), who gave us permission to use the videos and thus made this work possible. 

\section*{Limitations} \label{sec:limitations}
\vspace{-3mm}

%In this section, we discuss some of the limitations of \benchmarkName\  and corresponding challenges. % associated with it.

%\noindent \benchmarkName\ consists of content from the education and news domain. Hence, it is possibly prone to biases when it comes to developing generic sign language understanding systems. Consequently, we do not recommend using the iSign dataset for developing ML-based sign language models for real-world applications. The purpose of the benchmark proposed in this paper is to compare various technological developments made in the field of sign language processing, and help in drawing more research interest in sign language processing with a focus on transparency and reproducibility. We release the dataset via non-commercial research only license.

\noindent The large number of ISL-English sentence pairs in the initial version of the dataset makes it challenging to validate. Though we provide a small-scale validation with an ISL instructor's help, the entire corpus's validation is tedious, and it is infeasible to look into minute aspects of ISL-English translations in an initial version. In the future, we plan to extend the validation task and add a gold translation set with multiple references provided for a set of 5K ISL sentences. For an initial version of the benchmark, it is imperative to consider some possible technical constraints in the dataset. We believe addressing these constraints is a long-term goal, and mentioning them in detail will open up various directions for future work on analyzing and improving the benchmark.

\noindent\textbf{Alignment in ISL-English Pairs:} As the \benchmarkName\ translation dataset consists of sentence-level videos clipped from a longer video using multiple strategies, pauses in the available audio signal for ISLRTC videos, frame pattern heuristics for DEF videos, and available English caption timestamps for ISH videos, there are multiple ways in which the alignment between ISL signing sentence and corresponding English Sentence is disrupted. For example, in ISLRTC videos, though the audio in the background is aligned with the corresponding signs in the video, it could happen in a few cases that the audio was fast compared to the corresponding sign representation and may miss a few words at the beginning or the end of the sentence. %Similarly, for ISH videos, the timestamps in the captions might be delayed or fast in a few cases.

\noindent\textbf{\textbf{Co-referencing in ISL:}} As mentioned in Section \ref{sec:linguistics}, ISL linguistics involves assigning various short signs for names, places, etc., with various spatial reference points to refer to places defined for a specific conversation, making the context an essential feature for a complete translation of the content. As the translation dataset in \benchmarkName\ was created by segmenting a longer video into shorter segments representing a sentence, there is a high possibility that the names in the sentences are introduced in different sentences and assigned a short sign, which may be difficult to refer to in the later sentences. We consider this a major limitation of the created ISL-English pair dataset, as translating the same names would be difficult without a given reference to the created short sign or allocated space, and the independently made translations would result in a lower score in evaluation metrics like BLEU. One way of addressing this challenge would be to perform a task similar to NER on the ISL videos. However, the unavailability of resources and the scarcity of certified signers make ISL video annotation challenging. 

\noindent\textbf{Presence of Role Shifts:} As a major portion of the created dataset comes from educational content created by ISLRTC, a lot of material contains stories with fictional characters. In stories, there exists a high possibility of Role Shifts when producing sentences spoken by a character. A similar example in spoken storytelling would be the change in voice to articulate a character's voice, for example, the use of a squeaky voice for enacting the sentences of a small mouse. Since these Role Shifts are produced via non-manual markers, they would result in slight gesture variations for the same sentence, making the translated sentence more challenging to predict.

\noindent Lastly, in this paper, we do not compare ISL with other sign languages (like ASL and DGS); to the best of our knowledge, no such previous study exists, and performing such a study would require a considerable amount of effort in terms of humans having sign language expertise. In the future, we plan to explore such a study, as this can help with cross-model transfer, i.e., adapt models for rich resource sign languages to low resource sign languages.

\vspace{-3mm}
\section*{Ethical Considerations}
\vspace{-3mm}
We create a dataset from publicly available resources without violating copyright. We are not aware of any direct ethical concerns regarding our dataset. Moreover, the dataset involves people of Indian origin and is created mainly for Indian Sign Language translation. The ISL-specific insights are obtained from a professor working primarily on ISL linguistics. The annotations are done by the ISL professor and their team on a pro bono basis. 

\noindent \textbf{Please note we do not endorse the use of the benchmark data for non-research (commercial and real-life) applications, and the primary motivation for creating the \benchmarkName\ benchmark is to consolidate all the research happening in parallel for ISL.} %Hence we will release the benchmark and datasets under the Creative Common Attribution-NonCommercial-ShareAlike (CC BY-NC-SA) license. 
Moreover, we believe providing a platform by maintaining a common leaderboard for multiple tasks will advance the field with more transparency and reproducibility. Sign language datasets include visual modality with the inclusion of facial expression and non-manual
markers being an integral part of language, which does pose privacy challenges. Though the videos in the dataset are in the public domain, the models trained on the dataset may contain signer-specific features and may not generalize to real-world usage. %Hence, the primary usage of the benchmark is to promote research. % in natural language processing, gauge the state of progress and compare across different NLP techniques. %The benchmark does not advocate deploying any model trained on the dataset into the real world.

% \AJ{Official Review of Paper2324 by Reviewer jvV3: 
% However, the authors could further explore the potential negative societal impacts, such as privacy concerns and the risk of misinterpretation in critical communications.}

% % Bibliography entries for the entire Anthology, followed by custom entries
% %\bibliography{anthology,custom}
% % Custom bibliography entries only
% \bibliography{custom}
\bibliography{references}

\clearpage
\newpage

\appendix

\section*{Appendix}

\appendix

%%%%%%%%%%%%%%%%%%%%%%%%%%%

\titlecontents{section}[18pt]{\vspace{0.05em}}{\contentslabel{1.5em}}{}
{\titlerule*[0.5pc]{.}\contentspage} % Set the formatting for appendix sections in the table of contents

% % for list of tables
% \titlecontents{table}[0pt]{\vspace{0.05em}}{\contentslabel{1em}}{}
% {\titlerule*[0.5pc]{.}\contentspage} % Set the formatting for appendix tables in the list of tables

% for list of figures
\titlecontents{table}[0pt]{\vspace{0.05em}}{\contentslabel{1em}}{}
{\titlerule*[0.5pc]{.}\contentspage} % Set the formatting for appendix tables in the list of tables

\startcontents[appendix] % Start the table of contents for the appendix
\section*{Table of Contents} % Title for the appendix table of contents
%\addcontentsline{toc}{section}{Table of Contents} % Add the appendix table of contents to the main table of contents
\printcontents[appendix]{section}{0}{\setcounter{tocdepth}{4}} % Print the table of contents for the appendix

% \startlist[appendix]{lot} % Start the list of tables for the appendix
% \section*{List of Tables} % Title for the appendix list of tables
% %\addcontentsline{lot}{section}{List of Tables} % Add the appendix list of tables to the main list of tables
% \printlist[appendix]{lot}{}{\setcounter{tocdepth}{1}} % Print the list of tables for the appendix

\startlist[appendix]{lof} % Start the list of tables for the appendix
\section*{List of Figures} % Title for the appendix list of tables
%\addcontentsline{lot}{section}{List of Tables} % Add the appendix list of tables to the main list of tables
\printlist[appendix]{lof}{}{\setcounter{tocdepth}{1}} % Print the list of tables for the appendix

\newpage

%%%%%%%%%%%%%%%%%%%%%%%%%%%

%\section{iSign Benchmark Repository}

%%%%%%%%%%%%%%%%%%%%%%%%%%%

\section{iSign Dataset Details} \label{app:dataset-details}

% The dataset is made available at: \url{https://github.com/Exploration-Lab/iSign}
We plan to release the dataset along with a benchmark webpage to maintain the leaderboard of existing approaches to the proposed tasks.

\subsection{Dataset Creation Details}

To create the translation dataset with ISL-English video-sentence/phrase pairs, we primarily use three publicly available resources: ISLRTC videos,\footnote{\url{https://www.youtube.com/@islrtcnewdelhi4069/playlists}} ISH News (News channel in ISL),\footnote{\url{https://www.youtube.com/channel/UC99w_Bzj8ikOz8Gpv0prbNg}} and phrases from DEF (Deaf Enabled Foundation).\footnote{\url{https://www.youtube.com/channel/UCM7U7CyJGRIbu4qsmSh3UZg}}. These YouTube channels provide permission to scrape videos and use them for research. The scraped videos contained a signer communicating a sentence to the spectators. We obtain a list of $114$, $3373 $, and $651$ videos from ISLRTC, ISH, and DEF, respectively. These videos generally range from 15-20 minutes and contain about 20-25 number of sentences on average from various sources. To create a video-sentence/phrase level pair dataset, we further split the long videos at the sentence level. For ISH videos, the source videos provide an English caption aligned with the video where the same content is present in the English language as subtitles. We make use of these timestamps to split the videos into sign sentences and combine them with respective captions for English translation. For ISLRTC videos, since the subtitles were not present, we used the available English audio to generate the respective translations. At the end of each sentence, there is a pause in audio and the signer’s gesture; we clip the videos using an audio heuristic to create sentence-level segmentation of the signed video. For generating the respective transcripts, we use a speech-to-text model (Whisper \citep{Whisper}) for generating the text. Note that the speech-to-text model might not be $100\%$ accurate, resulting in some noisy texts for a few audio clips. Hence, we further clean the generated text via manual inspection by listening to the audio where the sentences are noisy and make less contextual sense. 
DEF videos provide a word-of-the-day format where a word is communicated at the start of the video with its explanation in the middle and some example sentences where the word is being used at the end. We clip all these sections to generate $3$ sets of clippings from a video, namely words, word descriptions, and examples. 
% \AJ{(Response to qzFm): Hence, we obtain the query candidate pairs for both tasks (Task 4 and Task 5) using the obtained $3$ sets of clippings from a video-specific video. Thank you for pointing this out. We will update the main paper with these details in the updated version of the paper, explicitly mentioning the process for creating the dataset in query candidate format.}
Note that a few videos have multiple example sentences available for the same word. We segment each example sentence as a different entry in the created dataset. 

\begin{figure}[h!]
\centering
  \includegraphics[width=\linewidth]{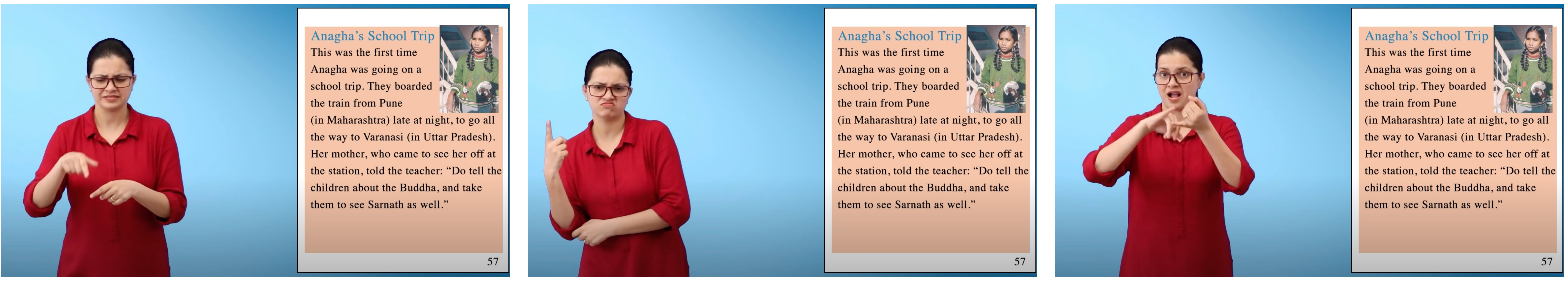}
  \caption{The figure shows an example of the educational content video where the signer signs for the corresponding textbook.}
  \label{fig:sample-video}
\end{figure}

\begin{figure}[h!]
\centering
  \includegraphics[width=\linewidth]{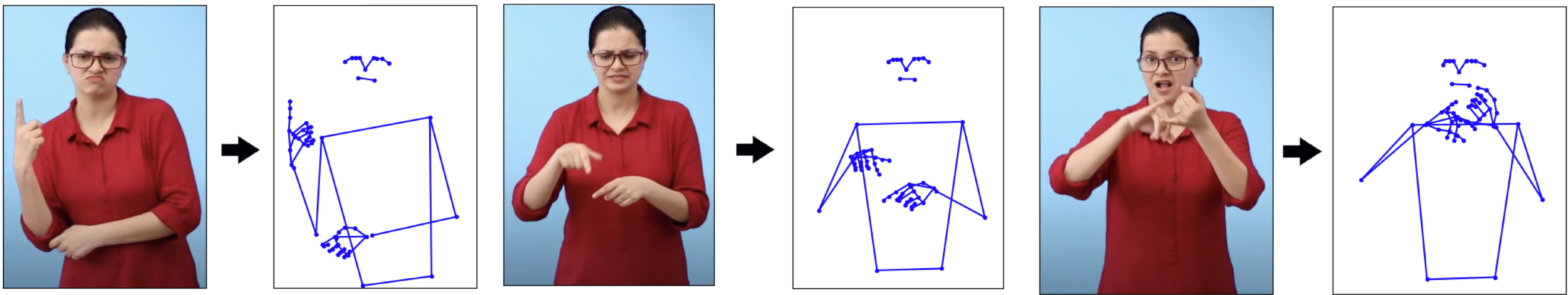}
  \caption{The figure shows an example of a frame from the iSign dataset video with the corresponding extracted keypoints.}
  \label{fig:sample-video-features}
\end{figure}

\begin{figure}[h!]
\centering
  \includegraphics[width=0.75\linewidth]{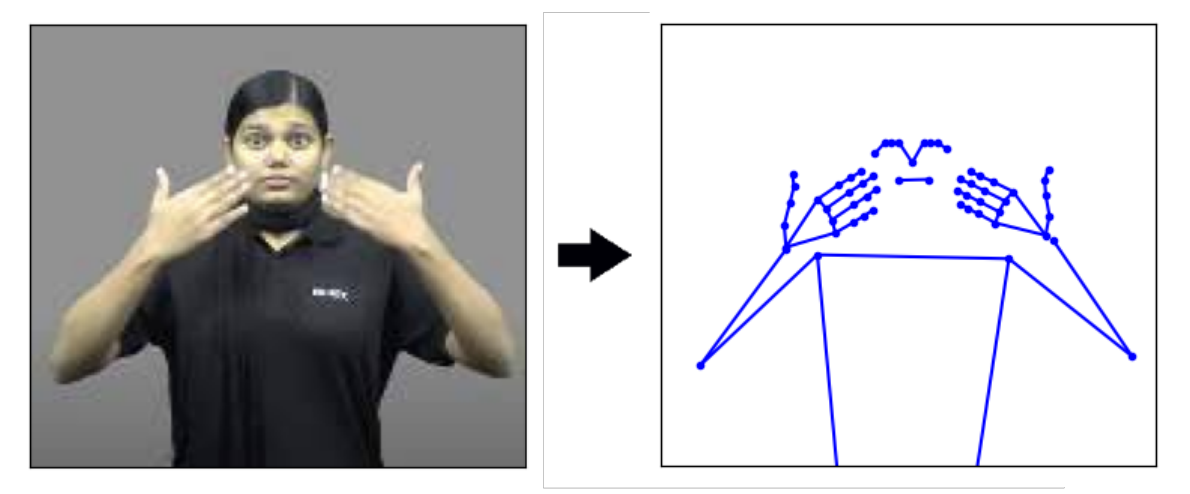}
  \caption{The figure shows an example of a frame from the CISLR dataset video with the corresponding extracted keypoints.}
  \label{fig:sample-video-features-cislr}
\end{figure}

\noindent\textbf{Data Cleaning and Preprocessing:} The videos (e.g., Fig. \ref{fig:sample-video}) contain the pictures corresponding book pages. We crop the signer out of the video by considering the face location of the first frame as the reference point and removing the remaining background in the videos. Further, the cropped videos are used to extract pose key points (Figure \ref{fig:sample-video-features} and Figure \ref{fig:sample-video-features-cislr}).

\begin{figure}[h!]
\centering
  \includegraphics[width=\linewidth]{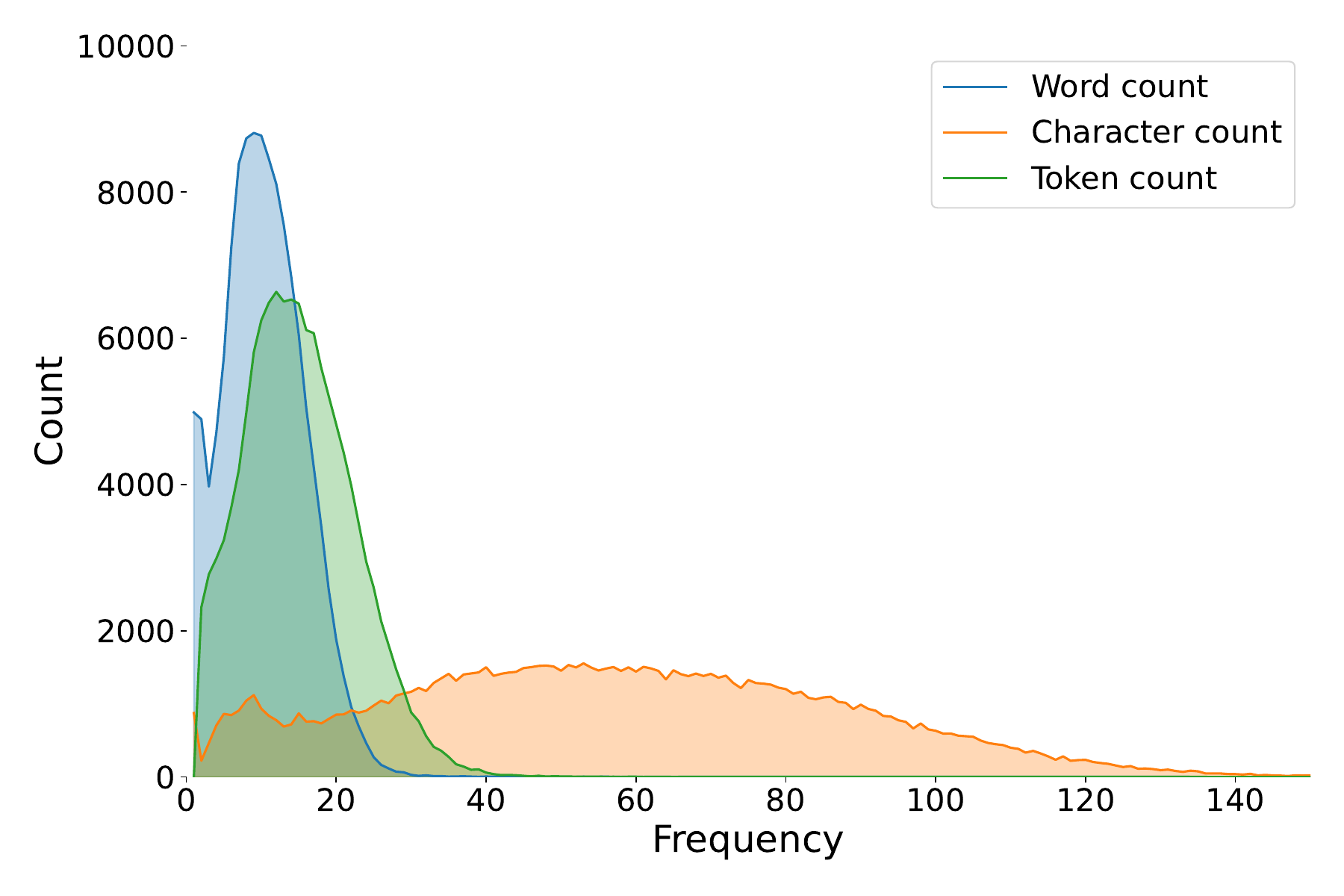}
  \caption{The figure shows the distribution of the number of words in the target translation text 
  % \AJ{this plot will change now after updating the dataset}
  }
  \label{fig:word-distribution}
\end{figure}

\noindent\textbf{Data Statistics:} We had a total of \iSignCount pairs of video-translation. We used T5-tokenizer \cite{huggingface-t5} for tokenizing the sentences, there were 15 tokens on an average for each translation text. For the words, we used `` "(space) as the separating character. More details about the distribution can be found in Figure \ref{fig:word-distribution} and Table \ref{app-tab:stats}.

\begin{table}[h!]
\centering
\tiny
\renewcommand{\arraystretch}{0.9} % Adjust the row height as needed
\begin{tabular}{@{}cc@{}}
\toprule
\begin{tabular}{@{}c@{}}   \textbf{Abbreviation} \end{tabular} & \textbf{Language Name} \\
\midrule
DGS & Deutsche Gebärdensprache \\
CSL & Chinese Sign Language \\
BSL & British Sign Language \\
ASL & American Sign Language \\
GSL & Ghanaian Sign Language \\
NSL & Nigerian Sign Language \\
KSL & Kenyan Sign Language \\
ZSL & Zambian Sign Language \\
ZISL & Zimbabwean Sign Language \\
SASL & South African Sign Language \\
ISL & Indian Sign Language \\
DSGS & Deutschschweizer Gebärdensprache (Swiss German Sign Language) \\
VGT & Vlaamse Gebarentaal (Flemish Sign Language) \\
\bottomrule
\end{tabular}
\caption{Sign Language Abbreviations and full forms corresponding to Table \ref{tab:translation-comparison}.}
\label{app-tab:sign-languages-fullform}
\end{table}

\begin{table}[h!]
\centering
\small
\setlength\tabcolsep{20pt}
\begin{tabular}{@{}lccc@{}}
\toprule
% \toprule
% & \textbf{Video-Sentence/Phrase Pairs (count = \iSignCount)}  \\ 
% & Stats & 
\midrule
 & Average & 15 \\
 & Minimum &  2\\
Tokens & Maximum &  146\\
 & Median &  14\\
 & $90^{th}$ percentile &  25\\
% \bottomrule

\midrule
 & Average &  10\\
 & Minimum &  1\\
Words & Maximum &  120\\
 & Median &  10\\
 & $90^{th}$ percentile & 17 \\
% \bottomrule

\midrule
 & Average &  57\\
 & Minimum &  1\\
Characters & Maximum &  622\\
 & Median &  56\\
 & $90^{th}$ percentile &  98\\
% \bottomrule

% \midrule
% &  \textbf{Video (count = \iSignCount)} & \\
\midrule
Frames & Average & 215 \\
& $90^{th}$ percentile & 371\\

\bottomrule
\end{tabular}
% \vspace{-5mm} % Adjust the spacing as needed
\caption{Statistics about the dataset: for the \iSignCount Video-Sentence/Phrase Pairs. The first four rows highlights the stats obtained from english phrase text sentence and the last row represents the stats for the corresponding Video-Sentence}
\label{app-tab:stats}
\end{table}

\begin{table*}[t]
\centering
\small
\renewcommand{\arraystretch}{1}
\setlength\tabcolsep{0.1pt}
\begin{tabular}{ccccccccc}
\toprule
\multicolumn{9}{c}{\textbf{Word Level Translation/Gloss Prediction}}                                                                                                                                                                                                 \\ \midrule
 \textbf{Task}                      & \textbf{Source Language} & \textbf{Target Language}        & 
 % \textbf{\# Test} & 
 \textbf{Model} & \textbf{Acc. (Top-1\%)} & \textbf{Acc. (Top-5\%)} & 
 % \textbf{Acc. (Top-10)} &
 \\ \midrule
ISLR & ISL                      & English/Gloss                   
% & 2285
&  I3D              & 10                          &   32.75  
\\
ISLR & ISL                      & English/Gloss                   
% & 2285
&  T5-small+I3D              & 15                         &   36.98  
\\
\toprule
\toprule
\multicolumn{9}{c}{\textbf{Sign Representation Learning}}                                                                                                                                                                                                                 \\ \midrule
\textbf{Task}                      & \textbf{Query}           & \textbf{Candidate}              & 
% \textbf{\# Test} & 
\textbf{Model} & \textbf{Top 5\% Acc.}         & 
% \textbf{Precision}        & \textbf{Recall} 
% & \textbf{F1}
% & 
\textbf{Rank (Avg.)}                \\ \midrule
          &            &  &  I3D                          &  $45$                &     $233/1523$                      &                           &                            &                                   \\ 
         &           &  &  T5-small+I3D                          &  $48$                &     $219/1523$                      &                           &                            &                                   \\ 
Word Presence           & word           & Ex. Sent. &  T5-small+Mediapipe(75)                          &  $42$                &     $244/1523$                      &                           &                            &                                   \\
        &         &  &  
T5-base + Mediapipe(75)                          &  $52$                &     $198/1523$                      &                           &                            &                                   \\
        &          &  &  
T5-large + Mediapipe(75)                          &  $43$                &     $237/1523$                      &                           &                            &                                   \\
\midrule
          &           &  &  I3D                          & $63$               & 59/593                           &                           &                            &                                    \\ 

         &           &  &  T5-small+I3D                          & $46$               & 169/593                           &                           &                            &                                    \\ 

Semantic Similarity           & word           & Descrip. Sent. &  T5-small+Mediapipe(75)                          & $47$               & 137/593                           &                           &                            &                                    \\ 

         &           &  &  T5-base+Mediapipe(75)                         & $67$               & 44/593                           &                           &                            &                                    \\ 

         &          &  &  T5-large+Mediapipe(75)                          & $53$               & 96/593                           &                           &                            &                                    \\ 
         \bottomrule
\bottomrule
\end{tabular}
\caption{Results of various baseline models on Task 3, 4 and 5. Ex. Sent. refers to Example Sentence and Descrip. Sent. refers to Description Sentence.}
\label{app-tab:results-task3-4-5}
\end{table*}

% \begin{enumerate}
%     \item Video Length vs. Token/word/character Lengths
%     \item Distribution of Train/Test/Validation splits (text sentence length, video length)
% \end{enumerate}

\subsection{Annotation Details} \label{app:annotation}
We asked three certified ISL instructors to translate and validate a random subset from the dataset (discussed in Section 3). One of the instructors is an assistant professor of sign language linguistics. All the instructors are employed with ISLRTC, the organization involved in creating the sign language content; however, the instructors did not participate in videos present in the translation dataset. The instructors performed the validation voluntarily. It took the instructor about 3 hours to validate 100 sentences. They generated the English translations by looking at the video. %We release the annotated set files along with the obtained comparison scores.

\section{Models and Experiment Details} \label{app:models-details}

\noindent\textbf{Pose Keypoint Extraction Pipeline: } We use the Mediapipe pose estimation pipeline.\footnote{\url{https://ai.googleblog.com/2020/12/mediapipe-holistic-simultaneous-face.html}} For the choice of holistic key points, we follow \citet{selvaraj-etal-2022-openhands}, which returns the 3D coordinates of 75 key points (excluding the face mesh). Figure \ref{fig:sample-video-features} and Figure \ref{fig:sample-video-features-cislr} shows an example of the obtained 75 keypoints from the mediapipe pipeline. Further, we normalize every frame’s key points by placing the midpoint of shoulder key points to the center and scaling the key points using the distance between the nose key point and the shoulders midpoint.

\noindent \textbf{Data Splits: } 
% Data split for Task-1 and Task-2  are shown in Table \ref{tab:train-val-test-split}. 
% Data split for Task-1 and Task-2  are shown in Table \ref{tab:train-val-test-split}. 
For Task-1 and Task-2, we use a split of 80\%, 10\%, and 10\% for train, validation, and test set, respectively. For Task-3, we follow CISLR \cite{joshi-etal-2022-cislr} and use 4765 samples as prototypes and the remaining 2285 videos as test sets. For Task-4 and Task-5, we take 594 and 1525 positive pairs, respectively to report the results.

% \begin{table}[h]
% \centering
% % \small
% % \renewcommand{\arraystretch}{1}
% % \setlength\tabcolsep{1pt}
% \begin{tabular}{cccc}
% \toprule
%           & Train & Validation & Test \\ \midrule
             
% \# Pairs & $\#$ 94582(80\%) & $\#$11824 (10\%)   & $\#$11822 (10\%) \\ 
% \bottomrule
% \end{tabular}
% \caption{The table shows the train, validation, and test split used for Task-1 and Task-2.}
% \label{tab:train-val-test-split}
% \end{table}

\noindent\textbf{Hyperparameters and Training:}
We follow the code base of SLT \cite{camgoz2020sign} to train and develop the proposed SLT-based pose-to-text architecture by modifying the input features to be sign-pose sequences generated by the mediapipe. The model architecture is a transformer-based encoder-decoder consisting of $3$ transformer layers each for both encoder and decoder. 
% The Pose-SLT baseline model has \AJ{update} trainable parameters and takes around  \AJ{30 minutes} to train for 1 epoch on the NVIDIA RTX 3090 GPU. 
We use the Adam optimizer \cite{kingma2014adam} 
with a learning rate of 0.0001, $\beta = (0.9, 0.999)$ and weight decay of  $0.0001$
for training the proposed baseline with a batch size of 32. The architecture has 14,337,264
trainable parameters 
% and takes around \AJ{$X$} minutes to train for 1 epoch on the NVIDIA A40 GPU
. For the generation task, we follow the code base released by \cite{saunders2020progressivetransformers}. The model architecture is a transformer-based encoder-decoder consisting of $2$ transformer layers for both the encoder and decoder. 
We use the Adam optimizer \cite{kingma2014adam} with a learning rate of 0.001 for training the proposed baseline with a batch size of 8. The architecture contains 3,67,68,768 trainable parameters.
We perform all the experiments using the NVIDIA A40 GPU machine.

\noindent We follow the T5 based model for sign language translation. We use the AdamW optimiser with a learning rate of 0.0001, and weight\_decay=$1e-6$. The batch size is 8 for a model with t5-large as backbone and 32 for a model with t5-small and base as backbones. We did transfer learning by freezing only the layers of T5 and finetune the embedding only upto 40 epochs, after that we train the whole model with reduced learning rate.

%\section{Discussion}

\section{Evaluation Metrics for Translation Tasks} \label{app:metrics}
In this section, we discuss some of the limitations of the current evaluation criteria for translation tasks and point toward the scope of building sign-language-specific evaluation methods in the future. We further provide the evaluation metric details of the additional tasks of Word Presence Prediction and Semantic Similarity Predictions.

\noindent\textbf{Sign Language to Spoken/Texual Language Translation:} In the current version of the benchmark, we use the standard translation metrics, which have shown effective usage and interpretation when working with written textual languages to textual language translations. However, in sign languages, the high use of co-referencing (explained in Section \ref{sec:linguistics}) makes the translations more challenging to divide into independent sentences. 
Current evaluation metrics like BLEU and ROUGE depend on the textual translation match between the reference-candidate pair translations. By taking a geometrical average over multiple values of $n$, BLEU scores give a measure to compute textual similarities between the reference-candidate pair translations. However, as the translated sentences depend on the co-reference symbol/gesture assigned in the past sentences of conversation, the exact independent generation for a sentence in a conversation becomes impossible. One way to overcome this issue is to perform a NER (Named-entity recognition) over the text translations to remove all the references to names and assign the same token for training the neural architectures. However, the co-reference usage for various things and concepts still needs to be solved. In the future, we plan to study and understand the co-reference feature of sign language in more detail, examining if a clear distinction or pattern exists between the usage of fingerspelling, new concepts, and location assigning for later use and how various tags can be introduced in the respective textual translations to make the sentences independent, facilitating the current neural translation pipelines working on language sentence pairs.

\noindent\textbf{Spoken/Texual Language to Sign Language Translation:}
Sign language generation is a more challenging task in terms of evaluation. In the current version of the benchmark, we use DTW scores to evaluate the quality of generated poses. Dynamic Time Warping, or DTW, tries to capture the similarity between two temporal sequences, takes care of the varying speed of the temporal sequences, and has shown practical usage in speech processing. However, for Indian Sign Language, the significant use of non-manual markers like facial
expressions, body language, etc., makes the DTW a less effective metric for capturing the quality of translations. One way of dealing with these would be to develop weighing criteria for various body keypoints associated with non-manual markers or detect specific action units for various gestures produced via non-manual markers. Due to a wide variety of gestures and sign production styles, designing a metric for judging the quality of translations becomes more challenging. In the past, \citet{saunders2020progressive} have proposed using back translation scores for judging the quality of generated keypoint poses. However, considering the low performance of machine translation systems for sign languages, the back-translation scores become less reliable. Moreover, a large-scale human evaluation study is needed to judge the effectiveness of various evaluation metrics, which currently needs to be added to the ISL literature. 

% \textcolor{green}
{\noindent\textbf{Word Presence Prediction:} In this task, we predict whether a specific query word is present in a candidate sign video. The primary evaluation metric for this task is the standard classification accuracy, which measures the model's ability to correctly predict the presence or absence of the query word within the candidate sentence. Additionally, since this task can be viewed as a retrieval task where the goal is to retrieve the correct candidate from a pool of candidates for a given query, another evaluation metric called Top Rank (Avg.) is considered. Top Rank (Avg.) involves ranking the entire pool of candidate sentences for a particular query and computing the average rank of the true candidate. A lower average rank signifies better performance, indicating that the model can retrieve the correct candidate more accurately from the pool.}

% \textcolor{green}
{
\noindent\textbf{Semantic Similarity Prediction:}
To evaluate the representations learned by the models for semantic similarity prediction, we use the top 5\% accuracy and rank metrics. The top 5\% accuracy determines if the true semantic match (the candidate video describing the query word) is ranked within the top 5\% of the similarity scores obtained for all candidate videos. Achieving a high Top 5\% accuracy indicates the model's effectiveness in accurately identifying the most semantically similar candidate videos for a given query.
The rank specifies the position of the true semantic match within the ranked list of candidate videos based on the similarity scores. A lower rank indicates that the true semantic match is positioned higher in the list, signifying a better performance of the model in identifying highly similar candidate videos for the query. }

\end{document}